\documentclass{article}
\usepackage{multirow}
\usepackage{amssymb}
\usepackage{latexsym}
\usepackage{amsmath}
\usepackage{subfigure}
\usepackage{booktabs}
\usepackage{url}
\usepackage{xcolor}
\definecolor{newcolor}{rgb}{.8,.349,.1}
\usepackage{algorithm}
\usepackage{algorithmic}
\usepackage{graphicx}
\usepackage{fullpage}
\newcommand{\POS}{\textsf{pos}}
\newcommand{\NEG}{\textsf{neg}}
\newcommand{\NEU}{\textsf{neu}}
\newcommand{\NONE}{\textsf{none}}
\newcommand{\MIX}{\textsf{mix}}

\begin{document}

\author{Eric S. Tellez\thanks{CONACyT Consejo Nacional de Ciencia y Tecnolog\'ia, Direcci\'on de C\'atedras, Insurgentes Sur 1582, Cr\'edito Constructor 03940, Ciudad de M\'exico, M\'exico.}~\thanks{INFOTEC Centro de Investigaci\'on e Innovaci\'on en Tecnolog\'ias de la Informaci\'on y Comunicaci\'on, Circuito Tecnopolo Sur No 112, Fracc. Tecnopolo Pocitos II, Aguascalientes 20313, M\'exico.} \\ \url{eric.tellez@infotec.mx}
\and Sabino Miranda-Jim\'enez\footnotemark[1]~\footnotemark[2] \\ \url{sabino.miranda@infotec.mx}
\and Mario Graff\footnotemark[1]~\footnotemark[2] \\ \url{mario.graff@infotec.mx}
\and Daniela Moctezuma\footnotemark[1]~\thanks{Centro de Investigaci\'on en Geograf\'ia y Geom\'atica ``Ing. Jorge L. Tamayo'', A.C. Circuito Tecnopolo Norte No. 117, Col. Tecnopolo Pocitos II, C.P. 20313,. Aguascalientes, Ags, M\'exico.}\\ \url{dmoctezuma@centrogeo.edu.mx}
\and Ranyart R. Su\'arez\thanks{Departamento de Estudios de Posgrado, Facultad de Ingenier\'ia El\'ectrica, Universidad Michoacana de San Nicol\'as de Hidalgo, Santiago Tapia 403, Morelia 58000, M\'exico} \\ \url{ranyart@dep.fie.umich.mx} \and Oscar S. Siordia\footnotemark[3] \\  \url{osanchez@centrogeo.edu.mx} }
\title{A Simple Approach to Multilingual Polarity Classification in Twitter}
\date{Sept. 2016}

\maketitle
\begin{abstract}
Recently, sentiment analysis has received a lot of attention due to the interest in mining opinions of social media users. Sentiment analysis consists in determining the polarity of a given text, i.e., its degree of positiveness or negativeness. Traditionally, Sentiment Analysis algorithms have been tailored to a specific language given the complexity of having a number of lexical variations and errors introduced by the people generating content. In this contribution, our aim is to provide a simple to implement and easy to use multilingual framework, that can serve as a baseline for sentiment analysis contests, and as starting point to build new sentiment analysis systems. We compare our approach in eight different languages, three of them have important international contests, namely, SemEval (English), TASS (Spanish), and SENTIPOLC (Italian). Within the competitions our approach reaches from medium to high positions in the rankings; whereas in the remaining languages our approach outperforms the reported results.

\end{abstract}




\section{Introduction}
\label{sec1}
Sentiment analysis is a crucial task in opinion mining field where the goal is to extract opinions, emotions, or attitudes to different entities (person, objects, news, among others). Clearly, this task is of interest for all languages; however, there exists a significant gap between English state-of-the-art methods and other languages. It is expected that some researchers decided to test the straightforward approach which consists in, first, translating the messages to English, and, then, use a high performing English sentiment classifier (for instance, see ~\cite{Balahur2012} and ~\cite{Balahur201456}) instead of creating a sentiment classifier optimized for a given language. However, the advantages of a properly tuned sentiment classifier have been studied for different languages (for instance, see \cite{arabic2016,arabic2015,Araujo2016,Bal2011}).

This manuscript focuses on the particular case of multilingual sentiment analysis of short informal texts such as Twitter messages. Our aim is to provide an easy-to-use tool to create sentiment classifiers based on supervised learning (i.e., labeled dataset) where the classifier should be competitive to those sentiment classifiers carefully tuned by some given languages. Furthermore, our second contribution is to create a well-performing baseline to compare new sentiment classifiers in a broad range of languages or to bootstrap new sentiment analysis systems. Our approach is based on selecting the best text-transforming techniques that optimize some performance measures where the chosen techniques are robust to typical writing errors.

In this context, we propose a robust multilingual sentiment analysis method, tested in eight different languages: Spanish, English, Italian, Arabic, German, Portuguese, Russian and Swedish. We compare our approach ranking in three international contests: TASS'15, SemEval'15-16 and SENTIPOLC'14, for Spanish, English and Italian respectively; the remaining languages are compared directly with the results reported in the literature. The experimental results locate our approach in good positions for all considered competitions; and excellent results in the other five languages tested. Finally, even when our method is almost cross-language, it can be extended to take advantage of language dependencies; we also provide experimental evidence of the advantages of using these language-dependent techniques.

The rest of the manuscript is organized as follows. Section~\ref{sec4} describes our proposed Sentiment Analysis method. Section~\ref{sec3} describes the datasets and contests used to test our approach; whereas, the experimental results, and, the discussion are presented on Section~\ref{section/experimental-results}. Finally, Section \ref{conclusions} concludes.

\section{Our Approach: Multilingual Polarity Classification}
\label{sec4}
We propose a method for multilingual polarity classification that can serve as a baseline as well as a framework to build more complex sentiment analysis systems due to its simplicity and availability as an open source software\footnote{https://github.com/INGEOTEC/b4msa}. As we mentioned, this baseline algorithm for multilingual Sentiment Analysis (B4MSA) was designed with the purpose of being multilingual and easy to implement. B4MSA is not a na\"ive baseline which is experimentally proved by evaluating it on several international competitions.

In a nutshell, B4MSA starts by applying text-transformations to the messages, then transformed text is represented in a vector space model (see Subsection~\ref{sec:preprocessing}), and finally, a Support Vector Machine (with linear kernel) is used as the classifier. B4MSA uses a number of text transformations that are categorized in cross-language features (see Subsection~\ref{sec:cross-lang}) and language dependent features (see Subsection~\ref{sec:lang-dependent}). It is important to note that, all the text-transformations considered are either simple to implement or there is a well-known library (e.g. \cite{NLTK2009,GENSIM}) to use them. It is important to note that to maintain the cross-language property, we limit ourselves to not use additional knowledge, this include knowledge from affective lexicons or models based on distributional semantics.

To obtain the best performance, one needs to select those text-transformations that work best for a particular dataset, therefore, B4MSA uses a simple random search and hill-climbing (see Subsection~\ref{section/parameter-optimization}) in space of text-transformations to free the user from this delicate and time-consuming task. Before going into the details of each text-transformation, Table~\ref{tab/parameters} gives a summary of the text-transformations used as well as their parameters associated.

\begin{table*}[!htb]
	\caption{Parameter list and a brief description of the functionality}
	\label{tab/parameters}
	\centering
	\resizebox{0.7\textwidth}{!}{
	\begin{tabular}{l@{~~}c@{~~}p{0.5\textwidth}} \hline
			\multicolumn{3}{c}{\bf cross-language features} \\ \hline
			\bf name & \bf values & \bf \hspace{3cm} description \\
			del-d1   & yes, no & If it is enabled then the sequences of repeated symbols are replaced by a single occurrence of the symbol. \\
			del-diac & yes, no & Determines if diacritic symbols, e.g., accent symbols, should be removed from the text. \\
			emo      & remove, group, none & Controls how emoticons are handled, i.e. removed, grouped by expressed emotion, or nothing. \\
			num      & remove, group, none & Controls how numbers are handled, i.e., removed, grouped into a special tag, or nothing. \\
			url      & remove, group, none & Controls how URLs are handled, i.e., removed, grouped into a special tag, or nothing. \\
			usr      & remove, group, none & Controls how users are handled, i.e., removed, grouped into a special tag, or nothing. \\
			lc       & yes, no & Letters are normalized to be lowercase if it is enabled \\

			tokenizer & $\mathcal{P}({\text{n-words} \cup \text{q-grams}})$ & One item among the power set (discarding the emptyset) of the union of *n-words and *q-grams. \\
			*n-words    & $\{1,2\}$      & The number of words used to describe a token. \\

			*q-grams  & $\{1,2,3,4,5,6,7\}$ & The length in characters of a token. \\
			\hline

			\multicolumn{3}{c}{\bf language dependent features} \\ \hline
			\bf name & \bf values & \bf \hspace{3cm} description \\

			stem   & yes, no & Determines if words are stemmed. \\

			neg    & yes, no & Determines if negation operators in the text are normalized
			and directly connected with the next content word. \\

			sw & remove, group, none & Controls how \textit{stopwords} are handled, i.e., removed, grouped, or left untouched. \\
		\hline
	\end{tabular}
	}
\end{table*}

\subsection{Cross-language Features}
\label{sec:cross-lang}
We defined cross-language features as a set of features that could be applied in most similar languages, not only related language families such as Germanic languages (English, German, etc.), Romance languages (Spanish, Italian, etc.), among others; but also similar surface features such as punctuation, diacritics, symbol duplication, case sensitivity, etc.
Later, the combination of these features will be explored to find the best configuration for a given classifier.

\subsubsection{Spelling Features}
Generally, Twitter messages are full of slang, misspelling, typographical and grammatical errors among others; in order to tackle these aspects we consider different parameters to study this effect. The following points are the parameters to be considered as spelling features. \textit{Punctuation} (\textit{del-punc}) considers the use of symbols such as question mark, period, exclamation point, commas, among other spelling marks. \textit{Diacritic symbols} (\textit{del-diac}) are commonly used in languages such as Spanish, Italian, Russian, etc., and its wrong usage is one of the main sources of orthographic errors in informal texts; this parameter considers the use or absence of diacritical marks. \textit{Symbol reduction} (\textit{del-d1}), usually, twitter messages use repeated characters to emphasize parts of the word to attract user's attention. This aspect makes the vocabulary explodes. We applied the strategy of replacing the repeated symbols by one occurrence of the symbol. \textit{Case sensitivity} (\textit{lc}) considers letters to be normalized in lowercase or to keep the original source; the aim is to cut the words that are the same in uppercase and lowercase.

\subsubsection{Emoticon (\textit{emo}) Feature}

We classified around 500 most popular emoticons, included text emoticons, and the whole set of unicode emoticons (around $1,600$) defined by~\cite{EmotUnicode2016} into three classes: positive, negative and neutral, which are grouped under its corresponding polarity word defined by the class name.

Table~\ref{table/emoticons} shows an excerpt of the dictionary that maps emoticons to their corresponding polarity class.

	\begin{table}
		\centering
		\begin{tabular}{|ccccc|}
			\hline
			  :)     &  :D     &  :P     & ~ $\rightarrow$ &	pos \\
			 :(      &  :-(    & :'(     & ~ $\rightarrow$ &	neg \\
			 :-$|$   &  U$\_$U & -.- & ~ $\rightarrow$ &	neu \\
            \hline
		\end{tabular}~\\~\\
		\caption{An excerpt of the mapping table from Emoticons to its polarity words.}
		\label{table/emoticons}
	\end{table}
\subsubsection{Word-based n-grams (\textit{n-words}) Feature}

N-words (word sequences) are widely used in many NLP tasks, and they have also been used in Sentiment Analysis~\cite{sidorov2012} and \cite{cui2015sentiment}. To compute the N-words, the text is tokenized and N-words are calculated from tokens. For example, let $T=\texttt{"the lights and shadows of your future"}$ be the text, so its 1-words (unigrams) are each word alone, and its 2-words (bigrams) set are the sequences of two words, the set ($W^T_2$), and so on. $W^T_2$ = \{\texttt{\small the lights, lights and, and shadows, shadows of, of your,\  your future}\}, so, given text of size $m$ words, we obtain a set containing at most $m-n+1$ elements. Generally, N-words are used up to 2 or 3-words because it is uncommon to find, between texts, good matches of word sequences greater than three or four words~\cite{Jurafsky2009}.

\subsubsection{Character-based q-grams (\textit{q-grams})}
In addition to the traditional N-words representation, we represent the resulting text as q-grams. A q-grams is an agnostic language transformation that consists in representing a document by all its substring of length $q$. For example, let $T=``\texttt{abra\_cadabra}"$ be the text, its 3-grams set are	\[Q^T_3 = \{ \texttt{abr, bra, ra\_, a\_c, \_ca, aca, cad, ada, dab}\},\] 	so, given text of size $m$ characters, we obtain a set with at most $m-q+1$ elements. Notice that this transformation handles white-spaces as part of the text. Since there will be q-grams connecting words, in some sense, applying q-grams to the entire text can capture part of the syntactic and contextual information in the sentence. The rationale of q-grams is also to tackle misspelled sentences from the approximate pattern matching perspective~\cite{NRbook02}.

\subsection{Language Dependent Features}
\label{sec:lang-dependent}
The following features are language dependent because they use specific information from the language concerned. Usually, the use of stopwords, stemming and negations are traditionally used in Sentiment Analysis. The users of this approach could add other features such as part of speech, affective lexicons, etc. to improve the performance \cite{BLiuBook2015}.

\subsubsection{Stopwords (\textit{sw}) Feature}

In many languages, there is a set of extremely common words such as determiners or conjunctions ($the$ or $and$) which help to build sentences but do not carry any meaning for themselves. These words are known as \textit{Stopwords}, and they are removed from text before any attempt to classify them. Generally, a stopword list is built using the most frequent terms from a huge document collection. We used the Spanish, English and Italian stopword lists included in the NLTK Python package~\cite{NLTK2009} in order to identify them.

\subsubsection{Stemming (\textit{stem}) Feature}

Stemming is a well-known heuristic process in Information Retrieval field that chops off the end of words and often includes the removal of derivational affixes. This technique uses the morphology of the language coded in a set of rules that are applied to find out word stems and reduce the vocabulary collapsing derivationally related words. In our study, we use the Snowball Stemmer for Spanish and Italian, and the Porter Stemmer for English that are implemented in NLTK package~\cite{NLTK2009}.

\subsubsection{Negation (\textit{neg}) Feature}

Negation markers might change the polarity of the message. Thus, we attached the negation clue to the nearest word, similar to the approaches used in \cite{sidorov2012}. A set of rules was designed for common negation structures that involve negation markers for Spanish, English and Italian. For instance, negation markers used for Spanish are {\em no} (not), {\em nunca, jam\'as} (never), and {\em sin} (without). The rules (regular expressions) are processed in order, and their purpose is to negate the nearest word to the negation marker using only the information on the text, e.g., avoiding mainly pronouns and articles. For example, in the sentence {\em El coche no es bonito} (The car is not nice), the negation marker {\em no} and {\em not} (for English) is attached to its adjective {\em no\_bonito} ({\em not\_nice}).

\subsection{Text Representation}
\label{sec:preprocessing}

After text-transformations, it is needed to represent the text in  suitable form in order to use a traditional classifier such as SVM. It was decided to select the well known vector representation of a text given its simplicity and powerful representation. Particularly, it is used the Term Frequency-Inverse Document Frequency which is a well-known weighting scheme in NLP. TF-IDF computes a weight that represents the importance of words or terms inside a document in a collection of documents, i.e., how frequently they appear across multiple documents. Therefore, common words such as \textit{the} and \textit{in}, which appear in many documents, will have a low score, and words that appear frequently in a single document will have high score. This weighting scheme selects the terms that represent a document.

\subsection{Parameter Optimization}
\label{section/parameter-optimization}

The model selection, sometimes called hyper-parameter optimization, is essential to ensure the performance of a sentiment classifier. In particular, our approach is highly parametric; in fact, we use such property to adapt to several languages. Table~\ref{tab/parameters} summarizes the parameters and their valid values. The search space contains more than 331 thousand configurations when limited to multilingual and language independent parameters; while the search space reaches close to 4 million configurations when we add our three language-dependent parameters. Depending on the size of the training set, each configuration needs several minutes on a commodity server to be evaluated; thus, an exhaustive exploration of the parameter space can be quite expensive making the approach useless in practice. To tackle the efficiency problems, we perform the model selection using two hyper-parameter optimization algorithms.

The first corresponds to {\em Random Search}, described in depth in \cite{bergstra2012random}. Random search consists on randomly sampling the parameter space and select the best configuration among the sample. The second algorithm consists on a {\em Hill Climbing}~\cite{burke2005search,battiti2008reactive} implemented with a memory to avoid testing a configuration twice. The main idea behind hill climbing {\em H+M} is to take a pivoting configuration, explore the configuration's neighborhood, and greedily moves to the best neighbor. The process is repeated until no improvement is possible. The configuration neighborhood is defined as the set of configurations such that these differ in just one parameter's value. This rule is strengthened for {\em tokenizer} (see Table~\ref{tab/parameters}) to differ in a single internal value not in the whole parameter value. More precisely, let $t$ be a valid value for {\em tokenizer} and $N(t)$ the set of valid values for neighborhoods of $t$, then $|t \cup s| \in \{|t|, |t| + 1\}$ and $|t \cap s| \in \{|t|, |t|-1\}$ for any $s \in N(t)$.

To guarantee a better or equal performance than random search, the {\em H+M} process starts with the best configuration found in the random search.
By using {\em H+M}, {\em sample size} can be set to 32 or 64, as rule of thumb, and even reach improvements in most cases (see \S\ref{section/experimental-results}).  Nonetheless, this simplification and performance boosting comes along with possible higher optimization times. Finally, the performance of each configuration is obtained using a cross-validation technique on the training data, and the metrics are usually used in classification such as: accuracy, score $F_1$, and recall, among others.

\section{Datasets and contests}
\label{sec3}

Nowadays, there are several international competitions related to text mining, which include diverse tasks such as: polarity classification (at different levels), subjectivity classification, entity detection, and iron detection, among others. These competitions are relevant to measure the potential of different proposed techniques. In this case, we focused on polarity classification task, hence, we developed a baseline method with an acceptable performance achieved in three different contests, namely, TASS'15 (Spanish)~\cite{villena2015overview}, SemEval'15-16 (English)~\cite{SemEval2015,SemEval2016}, and SENTIPOLC'14 (Italian)~\cite{SENTIPOLC2014}. In addition, our approach was tested with other languages (Arabic, German, Portuguese, Russian, and Swedish) to show that is feasible to use our framework as basis for building more complex sentiment analysis systems. From these languages, datasets and results can be seen in ~\cite{multilingual2016}, \cite{arabic2015} and \cite{arabic2016}.

Table~\ref{datasets} presents the details of each of the competitions considered as well as the other languages tested. It can be observed, from the table, the number of examples as well as the number of instances for each polarity level, namely, positive, neutral, negative and none. The training and development (only in SemEval) sets are used to train the sentiment classifier, and the gold set is used to test the classifier. In the case there dataset was not split in training and gold (Arabic, German, Portuguese, Russian, and Swedish) then a cross-validation (10 folds) technique is used to test the classifier. The performance of the classifier is presented using different metrics depending the competition. SemEval uses the average of score $F_{1}$ of positive and negative labels, TASS uses the accuracy and SENTIPOLC uses a custom metric (see~\cite{villena2015overview,SemEval2015,SemEval2016,SENTIPOLC2014}).

\begin{table}[!t]
 \begin{center}
  \caption{Datasets details from each competition tested in this work}
  \label{datasets}
  \resizebox{0.85\textwidth}{!}{
  \begin{tabular}{|c|l|r|r|r|r|r|}
   \cline{2-7}
   \multicolumn{1}{c|}{}  & \textbf{Dataset} &  \textbf{Positive} & \textbf{Neutral}  & \textbf{Negative} & \textbf{None} &\textbf{Total}\\
   \cline{1-7}
SemEval'15		& Training & 2,800 & 3,661 & 1,060 & - & 7,521\\ 
(English)   	& Development & 446 & 580 & 262 & - & 1,288 \\ 
				& Gold & 841 & 824 & 298 & - &1,963 \\ 
				\hline
SemEval'16		& Training & 3,094 & 2,043 & 863&-& 6,000 \\ 
(English)		& Development & 844 & 765 & 391 & -& 2,000 \\ 
				& Gold & 7,059 & 10,342 & 3,231 & -& 20,632 \\  
				\hline
TASS'15			& Training & 2,884 &  670 & 2,182& 1,482& 7,218 \\ 
(Spanish)		& Development & - & - & - & - & - \\ 
				& Gold 1K & 363 & 22 & 268 & 347&1,000 \\  
				& Gold 60K & 22,233 &  1,305 & 15,844&21,416 & 60,798 \\  \cline{2-7}
				\hline
SENTIPOL'14		& Training & 969 & 320 &  1,671 & 1,541 &4,501 \\ 
(Spanish)		& Development &- &- & - &-& - \\ 
				& Gold & 453 & 113 & 754 &607  &1,927 \\ 
				\hline
Arabic \cite{arabic2016,arabic2015}	& Unique & 448 & 202 & 1,350  & - &2,000 \\ \hline
German \cite{multilingual2016}	    & Unique & 23,860 & 50,368 & 17,274  & - & 91,502 \\ \hline
Portuguese \cite{multilingual2016}	& Unique & 24,595 & 29,357 & 32,110  & - & 86,062\\ \hline
Russian \cite{multilingual2016}	    & Unique & 19,238 & 28,665 & 21,197 & - & 69,100 \\ \hline
Swedish \cite{multilingual2016}	    & Unique & 13,265 & 15,410 & 20,580 & - & 49,255 \\
				\bottomrule

			\end{tabular}
            }
		\end{center}
	\end{table}

\section{Experimental Results}
\label{section/experimental-results}

We tested our framework on two kinds of datasets. On one hand, we compare our performance on three languages having well known sentiment analysis contests; here, we compare our work against competitors of those challenges. On the other hand, we selected five languages without popular opinion mining contests; for these languages, we compare our approach against research works reporting the used corpus.

\subsection{Performance on sentiment analysis contests}
\begin{figure*}[!htb]
\subfigure[SENTIPOLC'14]{
    \includegraphics[width=0.24\textwidth]{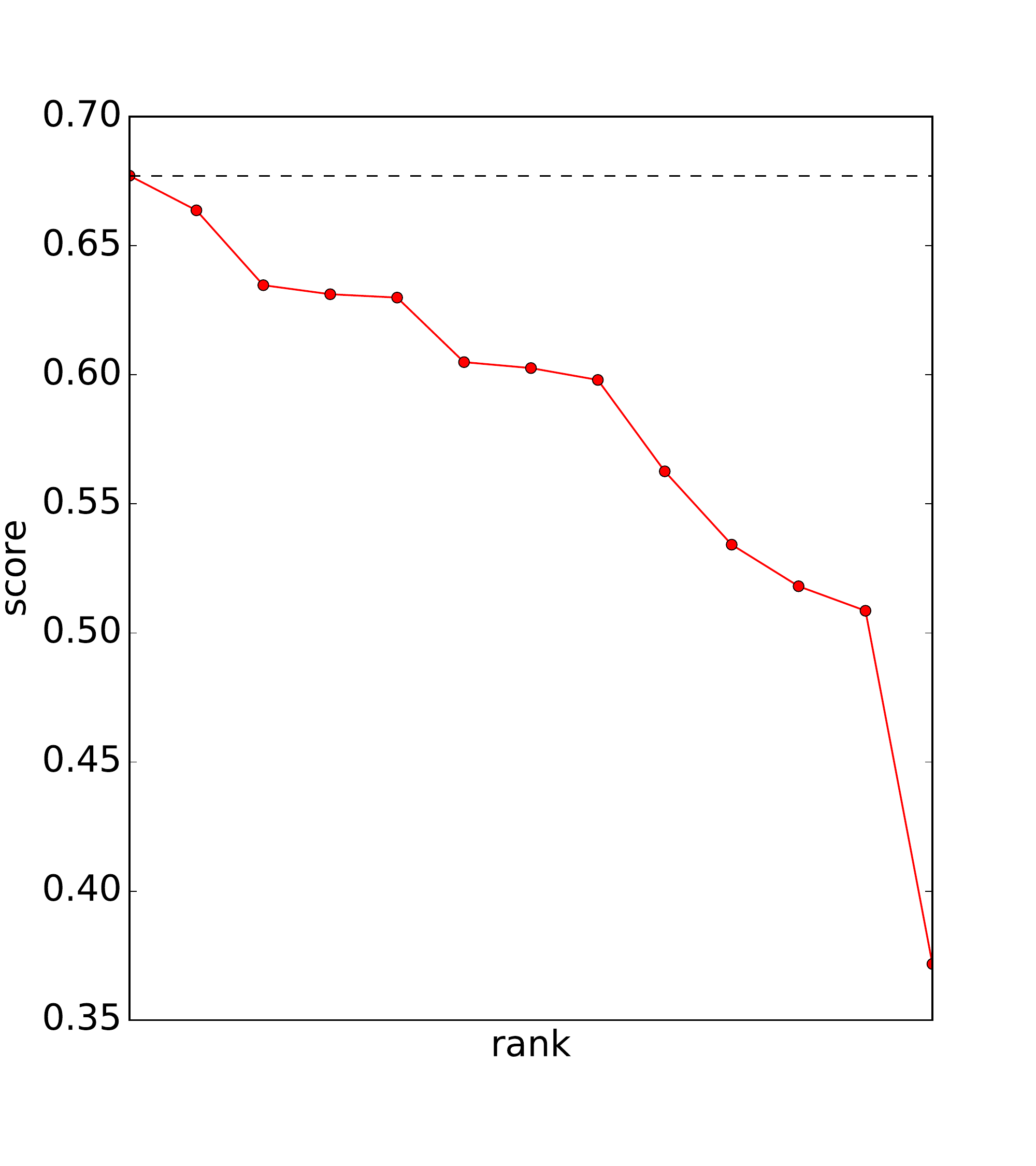}
    \label{fig/sentipolc14}
}\subfigure[TASS'15]{
\includegraphics[width=0.24\textwidth]{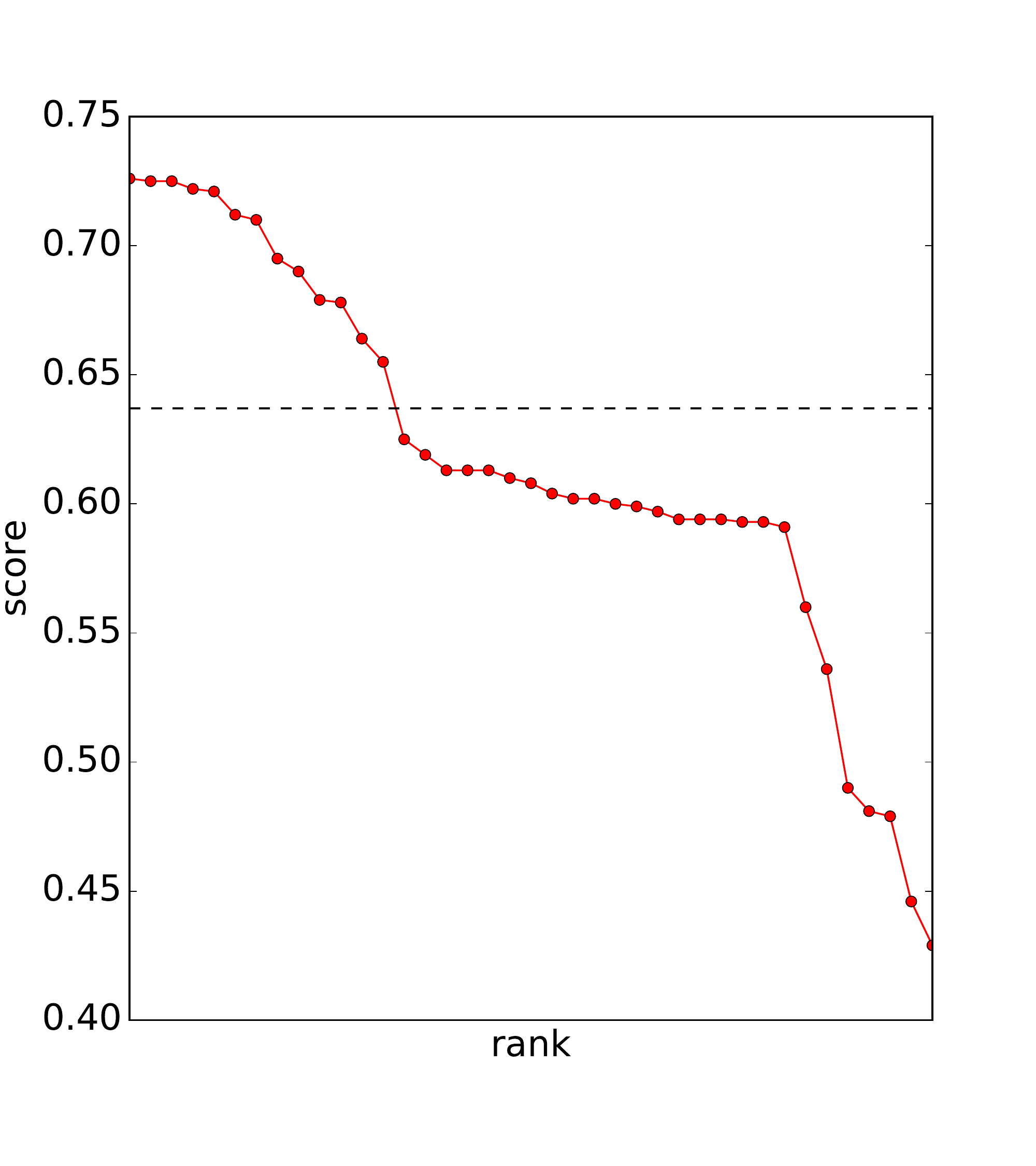}
\label{fig/tass15}
}\subfigure[SemEval'15]{
\includegraphics[width=0.24\textwidth]{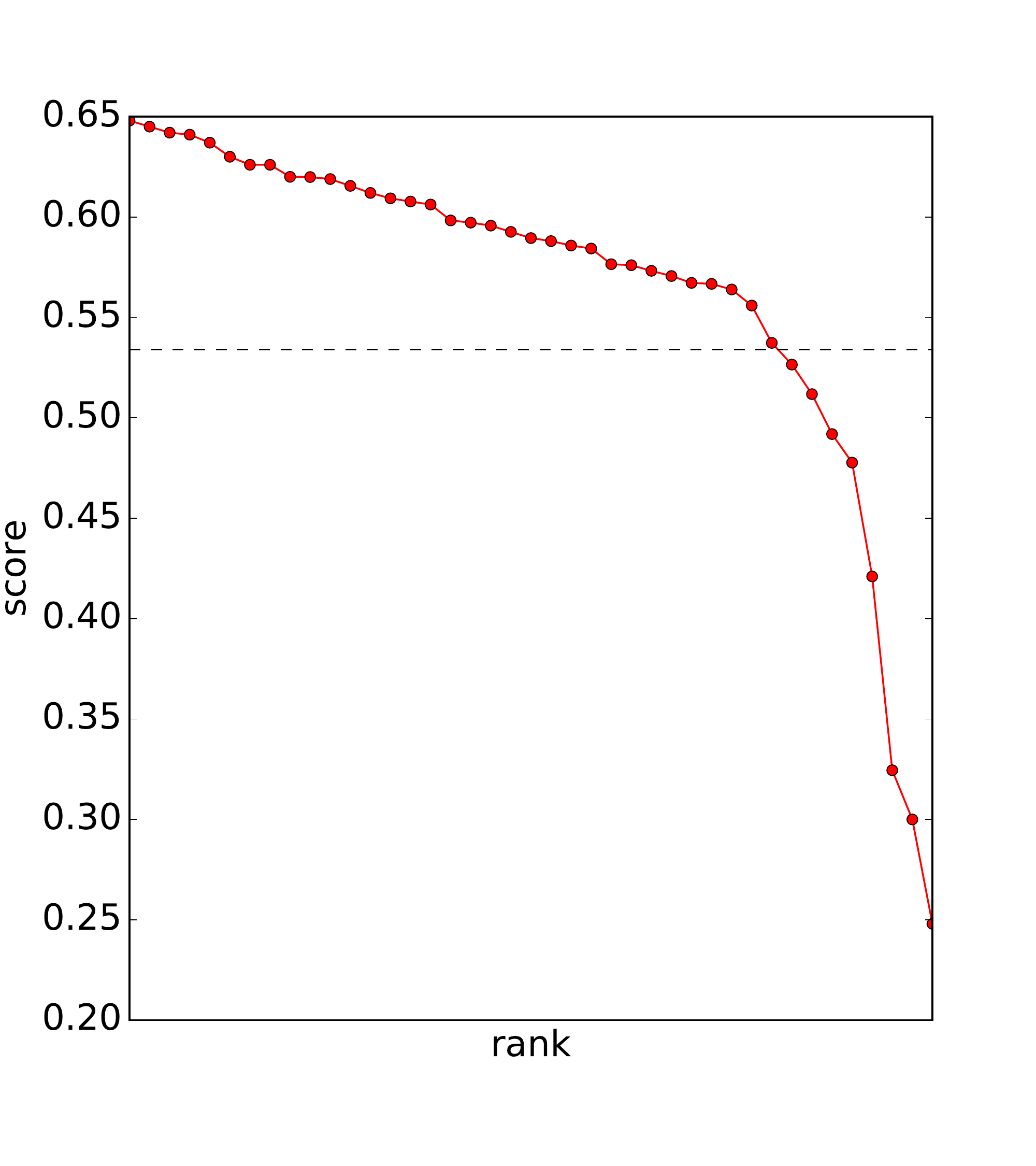}
\label{fig/semeval15}
}\subfigure[SemEval'16]{
\includegraphics[width=0.24\textwidth]{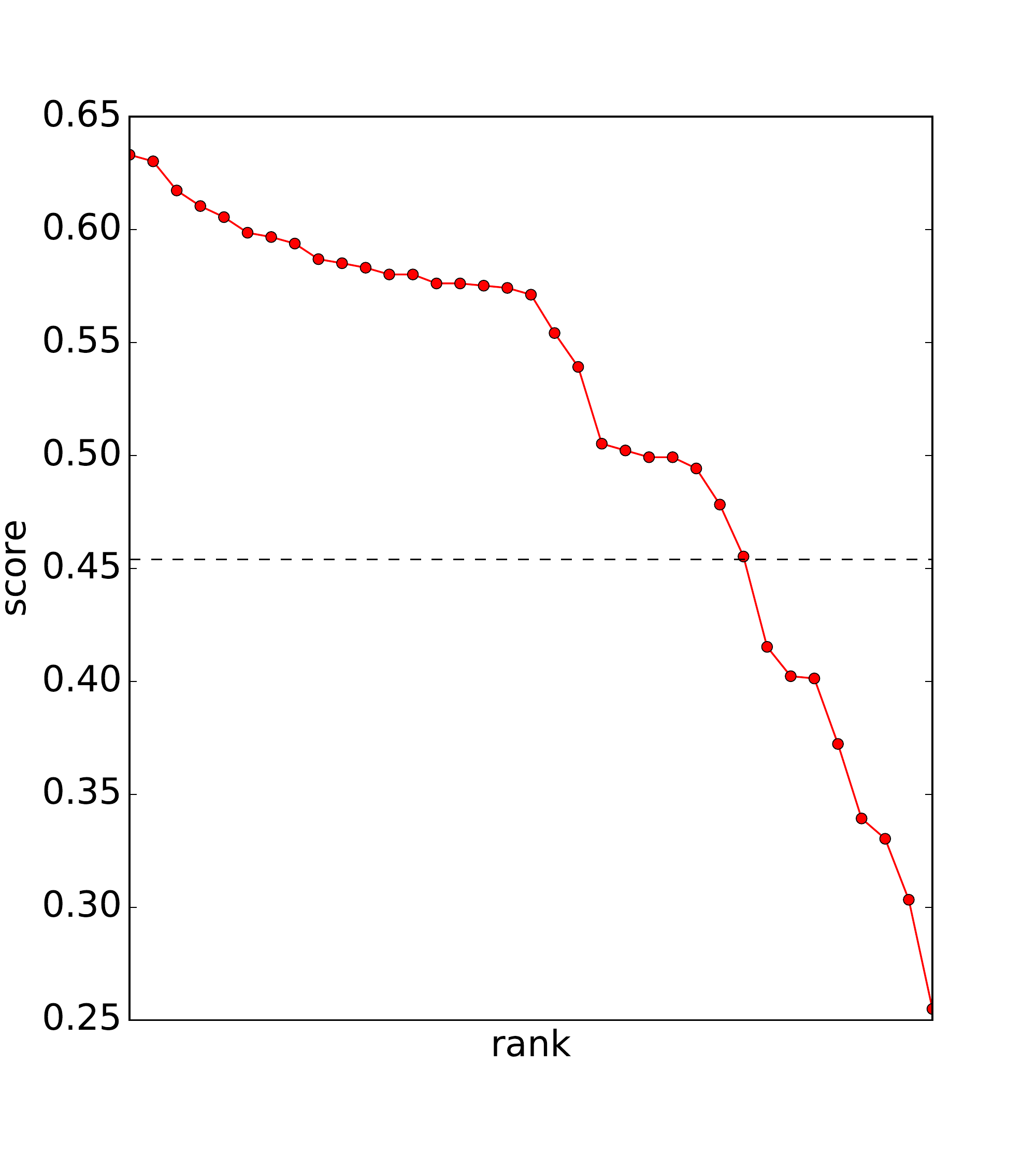}
\label{fig/semeval16}
}
\centering
\subfigure[B4MSA's performance summary; the last columns correspond to three standard scores.]{
	\centering\resizebox{0.9\textwidth}{!}{\begin{tabular}{r@{}lccccc|ccc}
			\toprule
			name & {} & language & classes & challenge's score &  rank  & score & acc. & macro $F_1$ & $\left(F^\POS_1 + F^\NEG_1\right)/2$\\
			\midrule
			SENTIPOLC&'14 & italian & $\{\POS,\NEG,\NONE, \MIX\}$ & custom (see~\cite{SENTIPOLC2014}) & 2 / 14 & 0.677 & 0.604 & 0.599 & 0.599 \\
			TASS&'15      & spanish & $\{\POS,\NEG,\NEU,\NONE\}$ & accuracy & 14 / 40 & 0.637 & 0.637 & 0.498 & 0.697\\
			SemEval&'15   & english & $\{\POS,\NEG,\NEU\}$ & $\left(F^\POS_1+F^\NEG_1\right)/2$ &    34 / 42 &  0.534 & 0.629 & 0.584 & 0.534 \\
			SemEval&'16   & english & $\{\POS,\NEG,\NEU\}$ & $\left(F^\POS_1+F^\NEG_1\right)/2$ &    28 / 36 &  0.454 & 0.534 & 0.477 & 0.454 \\
			\bottomrule
		\end{tabular}}
        \label{fig/summary}
}

\caption{
    The performance listing in four difference challenges. The horizontal lines appearing in a) to d) correspond to B4MSA's performance. All scores were computed using the official gold-standard and the proper score for each challenge.
    \label{fig/performance-b4msa}
}
\end{figure*}

Figure~\ref{fig/performance-b4msa} shows the performance on four contests, corresponding to three different languages. The performance corresponds to the multilingual set of features, i.e., we do not used language-dependent techniques.

Figures~\ref{fig/sentipolc14}-\ref{fig/semeval16} illustrates the results on each challenge, all competitors are ordered in score's descending order (higher is better). The achieved performance of our approach is marked with a horizontal line on each figure. Figure~\ref{fig/summary} briefly describes each challenge and summarizes our performance on each contest; also, we added three standard measures to simplify the insight's creation of the reader.

The winner method in SENTIPOLC'14 (Italian) is reported in ~\cite{UNIBA2014}. This method uses three groups of features: keyword and micro-blogging characteristics, Sentiment Lexicons, SentiWordNet and MultiWordNet, and Distributional Semantic Model (DSM) with a SVM classifier. In contrast with our method, in ~\cite{UNIBA2014} three external sentiment lexicons dictionaries were employed; that is, external information.

In TASS'15 (Spanish) competition, the winner reported method was ~\cite{Hurtado2015}, which proposed an adaptation based on a tokenizer of tweets {\em Tweetmotif}~\cite{exploratory}, Freeling~\cite{FreeLing30} as lemmatizer, entity detector, morphosyntactic labeler and a translation of the Afinn dictionary.
In contrast with our method, \cite{Hurtado2015} employs several complex and expensive tools.
In this task we reached the fourteenth position with an accuracy of $0.637$. Figure~\ref{fig/tass15} shows the B4MSA performance to be over two thirds of the competitors.

The remaining two contests correspond to the SemEval'15-16. The B4MSA performance in SemEval is depicted in Figures~\ref{fig/semeval15} and~\ref{fig/semeval16}; here, B4MSA does not perform as well as in other challenges, mainly because, contrary to other challenges, SemEval rules promotes the enrichment of the official training set. To be consistent with the rest of the experiments, B4MSA uses only the official training set. The results can be significantly improved using larger training datasets; for example, joining SemEval'13 and SemEval'16 training sets, we can reach $0.54$ for SemEval'16, which improves the B4MSA's performance (see Table~\ref{fig/performance-b4msa}).

In SemEval'15, the winner method is ~\cite{WebisSemeval2015}, which combines three approaches among the participants of SemEval'13, teams: NRC-Canada, GU-MLT-LT and KLUE, and from SemEval'14 the participant TeamX all of them employing external information. In SemEval'16, the winner method was ~\cite{SwissCheese2016} is composed with an ensemble of two subsystems based on convolutional neural networks, the first subsystem is created using 290 million tweets, and the second one is feeded with 150 million tweets. All these tweets were selected from a very large unlabeled dataset through distant supervision techniques.

Table~\ref{multiperfornace} shows the multilingual set of techniques and the set with language-dependent techniques; for each, we optimized the set of parameters through {\em Random Search} and $H+M$ (see Subsection~\ref{section/parameter-optimization}). The reached performance is reported using both cross-validation and the official gold-standard. Please notice how $H+M$ consistently reaches better performances, even on small sampling sizes. The sampling size is indicated with subscripts in Table~\ref{multiperfornace}. Note that, in SemEval challenges, the cross-validation performances are higher than those reached by evaluating the gold-standard, mainly because the gold-standard does not follow the distribution of training set. This can be understood because the rules of SemEval promote the use of external knowledge.

\begin{table*}[!h]
	\caption{B4MSA's performance on cross-validation and gold standard. The subscript at right of each score means for the random-search's parameter (sampling size) needed to find that value.}
    \label{multiperfornace}

    \centering
	\resizebox{0.85\textwidth}{!}{
	\begin{tabular}{r@{}l cc cc cc cc }
		\toprule
		\multirow{3}{*}{Dataset} & & \multicolumn{4}{c}{Multilingual Parameters} & \multicolumn{4}{c}{Language-Dependent Parameters}\\
		& &  \multicolumn{2}{c}{Random Search} &  \multicolumn{2}{c}{H+M Search} &  \multicolumn{2}{c}{Random Search} &  \multicolumn{2}{c}{H+M Search} \\

		& & cross-val. & gold-std. & cross-val. & gold-std. & cross-val. & gold-std. & cross-val. & gold-std. \\
		\midrule

		SENTIPOLC&'14 & - & $0.678_{256}$ & - & $0.677_{16}$ & - & $0.675_{8}$ & - & $0.674_{256}$\\
		TASS&'15 & $0.643_{128}$ & $0.636_{128}$ & $0.648_{8}$ & $0.637_{8}$ & $0.644_{256}$ & $0.635_{256}$ & $0.649_{32}$ & $0.637_{32}$\\
		SemEval&'15 & $0.585_{256}$ & $0.530_{256}$ & $0.590_{8}$ & $0.534_{8}$ & $0.590_{256}$ & $0.520_{256}$ & $0.596_{128}$ & $0.528_{128}$\\
		SemEval&'16 &
        $0.575_{64}$ & $0.456_{64}$ & $0.578_{64}$ & $0.454_{64}$ &
        $0.580_{256}$ & $0.462_{256}$ & $0.583_{256}$ & $0.462_{256}$\\

		\bottomrule
	\end{tabular}
	}
\end{table*}

Table~\ref{tab/multilingual-no-challenges} compares our performance on five different languages; we do not apply language-dependent techniques. For each comparison, we took a labeled corpus from \cite{arabic2015} (Arabic) and \cite{multilingual2016} (the remaining languages). According to author's reports, all tweets were manually labeled by native speakers as \POS, \NEG, or \NEU. The Arabic dataset contains $2,000$ items; the other datasets contain from 58 thousand tweets to more than 157 thousand tweets. We were able to fetch a fraction of the original datasets; so, we drop the necessary items to hold the original class-population ratio. The ratio of tweets in our training dataset, respect to the original dataset, is indicated beside the name. As before, we evaluate our algorithms through a 10-fold cross validation.

In \cite{arabic2015,arabic2016}, the authors study the effect of translation in sentiment classifiers; they found better to use native Arabic speakers as annotators than fine-tuned translators plus fine-tuned English sentiment classifiers. In \cite{multilingual2016}, the idea is to measure the effect of the agreement among annotators on the production of a sentiment-analysis corpus. Both, on the technical side, both papers use fine tuned classifiers plus a variety of pre-processing techniques to prove their claims. Table~\ref{tab/multilingual-no-challenges} supports the idea of choosing B4MSA as a bootstrapping sentiment classifier because, in the overall, B4MSA reaches superior performances regardless of the language. Our approach achieves those performance's levels since it optimizes a set of parameters carefully selected to work on a variety of languages and being robust to informal writing. The latter problem is not properly tackled in many cases.

\begin{table}[!h]
	\caption{Performance on multilingual sentiment analysis (not challenges). B4MSA was restricted to use only the multilingual set of parameters.}
	\label{tab/multilingual-no-challenges}
	\centering
    \resizebox{0.65\textwidth}{!} {
	\begin{tabular}{rlcccc}
		\toprule
		language       &  & $F_1$ & $\left(F^\POS_1 + F^\NEG_1\right)/2$ & acc \\
		\midrule \multirow{2}{*}{Arabic}
		& Salameh et al.~\cite{arabic2015}
		& -    & -     & 0.787 \\
		& Saif et al.~\cite{arabic2016}
		& -    & -     & 0.794 \\
		& B4MSA (100\%)   & 0.642 & 0.781  & 0.799 \\
		\midrule \multirow{2}{*}{German}
		& Mozeti{\v{c}} et al.~\cite{multilingual2016}
		& - & 0.536 & 0.610 \\
		& B4MSA (89\%)   & 0.621 & 0.559  & 0.668 \\
		\midrule \multirow{2}{*}{Portuguese}
		& Mozeti{\v{c}} et al.~\cite{multilingual2016}
		& -  & 0.553 & 0.507 \\
		& B4MSA (58\%)    & 0.550 & 0.591  & 0.555 \\
		\midrule \multirow{2}{*}{Russian}
		& Mozeti{\v{c}} et al.~\cite{multilingual2016}
		& -   & 0.615 & 0.603 \\
		& B4MSA (69\%)     & 0.754 & 0.768  & 0.750 \\
		\midrule \multirow{2}{*}{Swedish}
		& Mozeti{\v{c}} et al.~\cite{multilingual2016}
		& -   & 0.657 & 0.616 \\
		& B4MSA (93\%)     & 0.680 & 0.717  & 0.691 \\
		\bottomrule
	\end{tabular}
    }
\end{table}

\section{Conclusions}
\label{conclusions}

We presented a simple to implement multilingual framework for polarity classification whose main contributions are in two aspects. On one hand, our approach can serve as a baseline to compare other classification systems. It considers techniques for text representation such as spelling features, emoticons, word-based n-grams, character-based q-grams and language dependent features. On the other hand, our approach is a framework for practitioners or researchers looking for a bootstrapping sentiment classifier method in order to build more elaborated systems.

Besides the text-transformations, the proposed framework uses a SVM classifier (with linear kernel), and, hyper-parameter optimization using random search and {\em H+M} over the space of text-transformations. The experimental results show good overall performance in all international contests considered, and the best results in the other five languages tested.

It is important to note that all the methods that outperformed B4MSA in the sentiment analysis contests use extra knowledge (lexicons included) meanwhile B4MSA uses only the information provided by each contests. In future work, we will extend our methodology to include extra-knowledge in order to improve the performance.

\section*{Acknowledgements}
We would like to thank Valerio Basile, Julio Villena-Roman, and Preslav Nakov for kindly give us access to the gold-standards of SENTIPOLC'14, TASS'15 and SemEval 2015 \& 2016, respectively. The authors also thank Elio Villase\~nor for the helpful discussions in early stages of this research.

\bibliographystyle{plain}


\begin{thebibliography}{10}

\bibitem{Araujo2016}
Matheus Araujo, Julio Reis, Adriano Pereira, and Fabricio Benevenuto.
\newblock An evaluation of machine translation for multilingual sentence-level
  sentiment analysis.
\newblock In {\em Proceedings of the 31st Annual ACM Symposium on Applied
  Computing}, SAC '16, pages 1140--1145, New York, NY, USA, 2016. ACM.

\bibitem{Bal2011}
Daniella Bal, Malissa Bal, Arthur van Bunningen, Alexander Hogenboom, Frederik
  Hogenboom, and Flavius Frasincar.
\newblock {\em Sentiment Analysis with a Multilingual Pipeline}, pages
  129--142.
\newblock Springer Berlin Heidelberg, Berlin, Heidelberg, 2011.

\bibitem{Balahur2012}
Alexandra Balahur and Marco Turchi.
\newblock Multilingual sentiment analysis using machine translation?
\newblock In {\em Proceedings of the 3rd Workshop in Computational Approaches
  to Subjectivity and Sentiment Analysis}, WASSA '12, pages 52--60,
  Stroudsburg, PA, USA, 2012. Association for Computational Linguistics.

\bibitem{Balahur201456}
Alexandra Balahur and Marco Turchi.
\newblock Comparative experiments using supervised learning and machine
  translation for multilingual sentiment analysis.
\newblock {\em Computer Speech \& Language}, 28(1):56 -- 75, 2014.

\bibitem{UNIBA2014}
Pierpaolo Basile and Nicole Novielli.
\newblock Uniba at evalita 2014-sentipolc task: Predicting tweet sentiment
  polarity combining micro-blogging, lexicon and semantic features.
\newblock In {\em {Proceedings of the 4th evaluation campaign of Natural
  Language Processing and Speech tools for Italian (EVALITA'14)}}, Pisa, Italy,
  2014.

\bibitem{SENTIPOLC2014}
Valerio Basile, Andrea Bolioli, Malvina Nissim, Viviana Patti, and Paolo Rosso.
\newblock {Overview of the Evalita 2014 SENTIment POLarity Classification
  Task}.
\newblock In {\em {Proceedings of the 4th evaluation campaign of Natural
  Language Processing and Speech tools for Italian (EVALITA'14)}}, Pisa, Italy,
  2014.

\bibitem{battiti2008reactive}
Roberto Battiti, Mauro Brunato, and Franco Mascia.
\newblock {\em Reactive search and intelligent optimization}, volume~45.
\newblock Springer Science \& Business Media, 2008.

\bibitem{bergstra2012random}
James Bergstra and Yoshua Bengio.
\newblock Random search for hyper-parameter optimization.
\newblock {\em Journal of Machine Learning Research}, 13(Feb):281--305, 2012.

\bibitem{NLTK2009}
Steven Bird, Ewan Klein, and Edward Loper.
\newblock {\em {Natural Language Processing with Python}}.
\newblock O'Reilly Media, 2009.

\bibitem{burke2005search}
Edmund~K Burke, Graham Kendall, et~al.
\newblock {\em Search methodologies}.
\newblock Springer, 2005.

\bibitem{cui2015sentiment}
Zhijian Cui, Xiaodong Shi, and Yidong Chen.
\newblock Sentiment analysis via integrating distributed representations of
  variable-length word sequence.
\newblock {\em Neurocomputing}, 2015.

\bibitem{WebisSemeval2015}
Matthias Hagen, Martin Potthast, Michel Büchner, and Benno Stein.
\newblock Webis: an ensemble for twitter sentiment detection.
\newblock In {\em Proceedings of the 9th International Workshop on Semantic
  Evaluation (SemEval 2015}, pages 582--589. Association for Computational
  Linguistics, June 2015.

\bibitem{SwissCheese2016}
Jan Deriu; Maurice Gonzenbach; Fatih Uzdilli; Aurelien Lucchi; Valeria De
  Luca;~Martin Jaggi.
\newblock Swisscheese at semeval-2016 task 4: Sentiment classification using an
  ensemble of convolutional neural networks with distant supervision.
\newblock In {\em Proceedings of the 10th International Workshop on Semantic
  Evaluation}, SemEval '16, San Diego, California, June 2016. Association for
  Computational Linguistics.

\bibitem{Jurafsky2009}
Daniel Jurafsky and James~H. Martin.
\newblock {\em Speech and Language Processing (2Nd Edition)}.
\newblock Prentice-Hall, Inc., Upper Saddle River, NJ, USA, 2009.

\bibitem{exploratory}
Michel Krieger and David Ahn.
\newblock Tweetmotif: exploratory search and topic summarization for twitter.
\newblock In {\em In Proc. of AAAI Conference on Weblogs and Social}, 2010.

\bibitem{BLiuBook2015}
Bing Liu.
\newblock {\em Sentiment Analysis: Mining Opinions, Sentiments, and Emotions}.
\newblock Cambridge University Press, 2015.
\newblock ISBN: 1-107-01789-0. 381 pages.

\bibitem{Hurtado2015}
Ferran~Pla Llu\'is-F~Hurtado and Davide Buscaldi.
\newblock Elirf-upv en tass 2015: Análisis de sentimientos en twitter.
\newblock In {\em TASS 2015: Workshop on Sentiment Analysis at SEPLN co-located
  with 31st SEPLN Conference (SEPLN 2015)}, pages 75--79. Villena Rom{\'a}n,
  Julio; Garc{\'\i}a Morera, Janine; Garc{\'\i}a Cumbreras, Miguel {\'A}ngel;
  Mart{\'\i}nez C{\'a}mara, Eugenio; Mart{\'\i}n Valdivia, M. Teresa; Ure{\~n}a
  L{\'o}pez, L. Alfonso, 2015.

\bibitem{arabic2016}
Saif~M. Mohammad, Mohammad Salameh, and Svetlana Kiritchenko.
\newblock How translation alters sentiment.
\newblock {\em Journal of Artificial Intelligence Research}, 55:95--130, 2016.

\bibitem{multilingual2016}
Igor Mozeti{\v{c}}, Miha Gr{\v{c}}ar, and Jasmina Smailovi{\'c}.
\newblock Multilingual twitter sentiment classification: The role of human
  annotators.
\newblock {\em PloS one}, 11(5):e0155036, 2016.

\bibitem{SemEval2016}
Preslav Nakov, Alan Ritter, Sara Rosenthal, Veselin Stoyanov, and Fabrizio
  Sebastiani.
\newblock {SemEval}-2016 task 4: Sentiment analysis in {T}witter.
\newblock In {\em Proceedings of the 10th International Workshop on Semantic
  Evaluation}, SemEval '16, San Diego, California, June 2016. Association for
  Computational Linguistics.

\bibitem{NRbook02}
G.~Navarro and M.~Raffinot.
\newblock {\em Flexible Pattern Matching in Strings -- Practical on-line search
  algorithms for texts and biological sequences}.
\newblock Cambridge University Press, 2002.
\newblock ISBN 0-521-81307-7. 280 pages.

\bibitem{FreeLing30}
Lluís Padró and Evgeny Stanilovsky.
\newblock Freeling 3.0: Towards wider multilinguality.
\newblock In {\em Proceedings of the Language Resources and Evaluation
  Conference (LREC 2012)}, Istanbul, Turkey, May 2012. ELRA.

\bibitem{GENSIM}
Radim {\v R}eh{\r u}{\v r}ek and Petr Sojka.
\newblock {Software Framework for Topic Modelling with Large Corpora}.
\newblock In {\em {Proceedings of the LREC 2010 Workshop on New Challenges for
  NLP Frameworks}}, pages 45--50, Valletta, Malta, May 2010. ELRA.
\newblock \url{http://is.muni.cz/publication/884893/en}.

\bibitem{SemEval2015}
Sara Rosenthal, Preslav Nakov, Svetlana Kiritchenko, Saif Mohammad, Alan
  Ritter, and Veselin Stoyanov.
\newblock Semeval-2015 task 10: Sentiment analysis in twitter.
\newblock In {\em Proceedings of the 9th International Workshop on Semantic
  Evaluation (SemEval 2015)}, pages 451--463, Denver, Colorado, June 2015.
  Association for Computational Linguistics.

\bibitem{arabic2015}
Mohammad Salameh, Saif Mohammad, and Svetlana Kiritchenko.
\newblock Sentiment after translation: A case-study on arabic social media
  posts.
\newblock In {\em Proceedings of the 2015 Conference of the North American
  Chapter of the Association for Computational Linguistics: Human Language
  Technologies}, pages 767--777, Denver, Colorado, May--June 2015. Association
  for Computational Linguistics.

\bibitem{sidorov2012}
Grigori Sidorov, Sabino Miranda-Jim{\'e}nez, Francisco Viveros-Jim{\'e}nez,
  Alexander Gelbukh, No{\'e} Castro-S\'{a}nchez, Francisco Vel\'{a}squez,
  Ismael D\'{\i}az-Rangel, Sergio Su\'{a}rez-Guerra, Alejandro Trevi\~{n}o, and
  Juan Gordon.
\newblock Empirical study of machine learning based approach for opinion mining
  in tweets.
\newblock In {\em Proceedings of the 11th Mexican International Conference on
  Advances in Artificial Intelligence - Volume Part I}, MICAI'12, pages 1--14,
  Berlin, Heidelberg, 2013. Springer-Verlag.

\bibitem{EmotUnicode2016}
Unicode.
\newblock Unicode emoji chart.
\newblock \url{http://unicode.org/emoji/charts/full-emoji-list.html}, 2016.
\newblock Accessed 20-May-2016.

\bibitem{villena2015overview}
Julio Villena~Rom{\'a}n, Janine Garc{\'\i}a~Morera, Miguel~{\'A}ngel
  Garc{\'\i}a~Cumbreras, Eugenio Mart{\'\i}nez~C{\'a}mara, M.~Teresa
  Mart{\'\i}n~Valdivia, and L.~Alfonso Ure{\~n}a~L{\'o}pez.
\newblock Overview of tass 2015.
\newblock In {\em TASS 2015: Workshop on Sentiment Analysis at SEPLN co-located
  with 31st SEPLN Conference (SEPLN 2015)}, pages 13--21. Villena Rom{\'a}n,
  Julio; Garc{\'\i}a Morera, Janine; Garc{\'\i}a Cumbreras, Miguel {\'A}ngel;
  Mart{\'\i}nez C{\'a}mara, Eugenio; Mart{\'\i}n Valdivia, M. Teresa; Ure{\~n}a
  L{\'o}pez, L. Alfonso, 2015.

\end{thebibliography}

\end{document}